\documentclass[conference]{IEEEtran}
\IEEEoverridecommandlockouts

\usepackage{cite}
\usepackage{amsmath,amssymb,amsfonts}
\usepackage{graphicx}
\usepackage{subcaption}
\usepackage{booktabs}
\usepackage{multirow}
\usepackage{algorithm}
\usepackage{algpseudocode}
\usepackage{url}
\usepackage[hidelinks]{hyperref}

\begin{document}

\title{Patch-Based 3D Variational Autoencoder for Super-Resolution\\
of Turbulent Channel Flow}

\author{
\IEEEauthorblockN{Anuraj Maurya}
\IEEEauthorblockA{Indian Institute of Science, Bangalore, India\\
anurajmaurya@iisc.ac.in}
}

\maketitle

\begin{abstract}
Direct numerical simulation (DNS) of wall-bounded turbulence resolves the
full range of spatio-temporal scales, but at a cost that grows steeply with
Reynolds number and is prohibitive for routine engineering use.
Super-resolution offers a cheaper route: recover fine-scale structure from
an affordable coarse field. Most published super-resolution studies for
turbulence operate on two-dimensional sections, where vortex stretching is
absent by construction, and extend poorly to three dimensions because the
parameter count grows with the volume being reconstructed. We present a
three-dimensional variational autoencoder (3D-VAE) that avoids this scaling
by reconstructing a local $16^3$ block from a coarse block spanning a larger
physical extent, then applying the learned operator convolutionally across
the domain with overlap averaging. The parameter count therefore depends on
the patch size rather than the domain size. Training uses streamwise
velocity from a single snapshot of DNS of turbulent channel flow at
$Re_\tau \approx 1000$ from the Johns Hopkins Turbulence Database, with
evaluation on a held-out snapshot. Against DNS the 3D-VAE attains a mean
absolute error of $0.055$, compared with $0.075$ for tricubic and $0.076$
for Lanczos interpolation, and reduces the mean absolute error of the
two-dimensional spectral amplitude by approximately a factor of three
($0.91$ against $2.63$ and $2.85$). Applied to coarse finite-element
simulations of the same configuration, the same architecture recovers
spectral content absent from its input. A conditional 3D-GAN trained on
identical data failed to converge under Wasserstein training and is
reported as a negative result. We identify the principal failure
modes---damping of the smallest resolved scales, periodic artefacts at the
patch stride, and under-prediction of the tails of the velocity
distribution---and state the conditions under which the results were
obtained.
\end{abstract}


\section{Introduction}

The Navier--Stokes equations describe fluid motion completely, but admit no
general analytical solution in the turbulent regime. Direct numerical
simulation (DNS) resolves all dynamically relevant scales and is treated as
ground truth, yet its cost grows so steeply with Reynolds number that it
remains inaccessible for most engineering
geometries~\cite{pope2000turbulent}. Practical computation therefore relies
on closure: Reynolds-averaged Navier--Stokes models represent the Reynolds
stress in terms of mean quantities, while large-eddy simulation (LES)
resolves the energetic scales and models the subgrid-scale (SGS)
contribution. Both introduce modelling error precisely where the
small-scale physics matters, and the classical closures---eddy-viscosity
two-equation models, one-equation transport models such as
Spalart--Allmaras~\cite{spalart1992one}, and variational multiscale
formulations~\cite{hughes1998variational}---are known to degrade under
strong pressure gradients, curvature, and near-wall anisotropy.

Data-driven closure has become an active alternative, beginning with Milano
and Koumoutsakos~\cite{milano2002neural} on near-wall reconstruction and
expanding with deep learning: spatially varying corrections to turbulence
models~\cite{zhang2015machine}, Reynolds stress anisotropy with
architecturally embedded Galilean invariance~\cite{ling2016reynolds}, SGS
reaction rates~\cite{lapeyre2019training}, and synthetic inflow
generation~\cite{fukami2019synthetic}. Reviews by
Kutz~\cite{kutz2017deep}, Duraisamy \emph{et
al.}~\cite{duraisamy2019turbulence}, Brunton \emph{et
al.}~\cite{brunton2020machine}, and Vinuesa and
Brunton~\cite{vinuesa2022enhancing} survey the space.

A distinct but related problem is \emph{super-resolution} (SR): rather than
modelling an unclosed term, learn a map from a coarse velocity field to a
fine one. Fukami \emph{et al.} established that convolutional networks can
recover DNS-resolution two-dimensional fields from grossly under-resolved
input~\cite{fukami2019super}, and the survey by Fukami \emph{et
al.}~\cite{fukami2023survey} catalogues the methods that followed:
GAN-based~\cite{ledig2017photo,yousif2021esrgan,deng2019super},
physics-constrained~\cite{gao2021pinn,ren2023physr},
transformer-based~\cite{xu2023transformer}, and
diffusion-based~\cite{shu2023diffusion,shan2024pird} formulations.

Two gaps motivate this work. First, the majority of these studies are
two-dimensional. Vortex stretching---the mechanism transferring energy to
small scales---has no two-dimensional analogue, so two-dimensional results
do not transfer to three dimensions by default; the three-dimensional
studies that exist typically target modest upsampling factors or small
domains~\cite{xu2020data,subramaniamturbulence,trinh2024vehicle}. Second,
and more restrictive: a network mapping an entire coarse domain to an
entire fine domain has a parameter count that grows with domain size, which
caps the achievable domain for a fixed memory budget.

This paper addresses the second gap. We hypothesise that a coarse
representation over a spatial neighbourhood carries sufficient information
to reconstruct the fine-scale field at the \emph{centre} of that
neighbourhood, and that the resulting local map can be applied
convolutionally across an arbitrarily large domain. Our contributions are:

\begin{itemize}
  \item A patch-based 3D-VAE for turbulence super-resolution whose cost is
        decoupled from domain size, with a sampling scheme
        (Sec.~\ref{sec:sampling}) that generates training data from a single
        DNS snapshot.
  \item Quantitative comparison against tricubic and Lanczos interpolation
        in both physical and spectral space (Sec.~\ref{sec:results}),
        showing a $27\%$ reduction in mean absolute error and an
        approximately threefold reduction in spectral amplitude error.
  \item Demonstration that the DNS-trained model transfers to coarse
        finite-element fields produced by a different numerical scheme
        (Sec.~\ref{sec:les2dns}).
  \item A reported negative result for a conditional 3D-GAN on identical
        data (Sec.~\ref{sec:gan}), and an explicit account of failure modes
        and reproducibility conditions (Sec.~\ref{sec:limitations}).
\end{itemize}

\section{Related Work}

\subsection{CNN-based super-resolution}
Flow-field SR inherited its architectures from image processing: SRCNN and
its accelerated variant~\cite{dong2015image,dong2016accelerating}, deep
recursive residual networks~\cite{tai2017image}, and enhanced residual
networks~\cite{lim2017enhanced}. Fukami \emph{et
al.}~\cite{fukami2019super} adapted SRCNN to two-dimensional turbulence and
later extended it to joint spatio-temporal
reconstruction~\cite{fukami2021spatiotemporal}. Liu \emph{et
al.}~\cite{liu2020deep} compared static and multiple-temporal-path CNNs;
Bi \emph{et al.}~\cite{bi2022flowsrnet} added multi-scale integration;
G\"uemes \emph{et al.}~\cite{guemes2021coarse} reconstructed velocity
fields from wall measurements alone. Transfer learning has been used to
reduce the data requirement~\cite{obiols2021surfnet}. The common limitation
is that plain convolutional regression under an $L_2$ loss smooths
high-wavenumber content, which motivated the adversarial and generative
formulations below.

\subsection{GAN-based super-resolution}
Generative adversarial networks~\cite{goodfellow2014generative} entered
flow SR through SRGAN~\cite{ledig2017photo}. Deng \emph{et
al.}~\cite{deng2019super} applied SRGAN and ESRGAN to experimental
two-dimensional wake fields, taking streamwise and spanwise velocity
together with velocity magnitude as input channels and recovering $4\times$
and $8\times$ upsampling. Yousif \emph{et al.}~\cite{yousif2021esrgan}
extended enhanced-SRGAN to spatially limited turbulence data and showed
generalisation across Reynolds numbers unseen in training. In three
dimensions, Xu \emph{et al.}~\cite{xu2020data} upsampled LES of a piloted
CH$_4$/air diffusion flame by a factor of two in each direction using 3D
convolutions with skip connections, outperforming tricubic interpolation.
Subramaniam \emph{et al.}~\cite{subramaniamturbulence} introduced TEGAN and
TEResNet for $64^3$ forced isotropic turbulence downsampled to $16^3$,
adding continuity and pressure-Poisson residuals to the generator loss.
Macart and Attili~\cite{macart2024adversarial} isolated the effect of
adversarial training itself on reconstruction quality, finding that the
adversarial term improves high-wavenumber content but can degrade pointwise
accuracy---a trade-off directly relevant to the negative result in
Sec.~\ref{sec:gan}.

\subsection{Physics-constrained and operator-based approaches}
Gao \emph{et al.}~\cite{gao2021pinn} performed SR and denoising without
high-resolution labels by enforcing the governing equations directly. Ren
\emph{et al.}~\cite{ren2023physr} extended this to spatio-temporal data.
Kim \emph{et al.}~\cite{kim2021unsupervised} demonstrated unsupervised SR
of turbulence using cycle-consistency, removing the need for paired
coarse--fine data. Hassanaly \emph{et al.}~\cite{hassanaly2022adversarial}
framed SR as sampling from a high-dimensional conditional distribution,
which is the framing closest in spirit to the generative formulation used
here. Neural-operator and generative-model treatments have since been
applied to three-dimensional isotropic turbulence for joint
super-resolution, forecasting, and sparse
reconstruction~\cite{oommen2025generative}, and the same machinery has been
carried into turbulent reacting flow~\cite{pang2024reacting}.
Page~\cite{page2025dynamics} places the governing dynamics directly in the
training loss rather than as a soft residual penalty, which is arguably the
strongest current answer to the small-scale damping we report in
Sec.~\ref{sec:limitations}.

\subsection{Sampling strategy and the closest prior work}
The component of our method most in need of positioning is the patch-based
sampling scheme. Fukami and Taira~\cite{fukami2024single} showed---for both
two-dimensional isotropic turbulence and three-dimensional turbulent
channel flow---that a carefully designed model trained on flow tiles drawn
from a \emph{single} snapshot can reconstruct vortical structures, and
attributed this to the scale-invariance properties of turbulence. Their
work postdates the experiments reported here but is the more thorough
treatment of the underlying question, and we cite it as the strongest
available justification for single-snapshot patch training rather than as a
baseline we outperform. Related sampling and generalisation techniques are
discussed by Morimoto \emph{et al.}~\cite{morimoto2022generalization}.
Invariance-aware SGS modelling in a local
frame~\cite{prakash2022invariant} provides the formal notion of invariance
that a future version of this method should target.

\subsection{Variational multiscale learning}
Pradhan and Duraisamy~\cite{pradhan2021variational} take an approach
complementary to ours: instead of learning a generic coarse-to-fine map,
they derive the exact fine-scale solution for a one-dimensional
convection--diffusion problem, show that its $L_2$-optimal projection onto
a discontinuous basis depends only on the element P\'eclet number $\alpha$
and the normalised coarse-scale coefficients, and learn that
low-dimensional function. Their variational super-resolution network
multiplies separately embedded physics and solution features,
$u' = f^{\text{FNN}}(g_\alpha \odot g_u)$, so that the learned closure
inherits the structure of the variational multiscale formulation. Our
method makes no such structural guarantee; we return to this in
Sec.~\ref{sec:future}.

\section{Method}

\begin{figure*}[t]
\centering
\includegraphics[width=\linewidth]{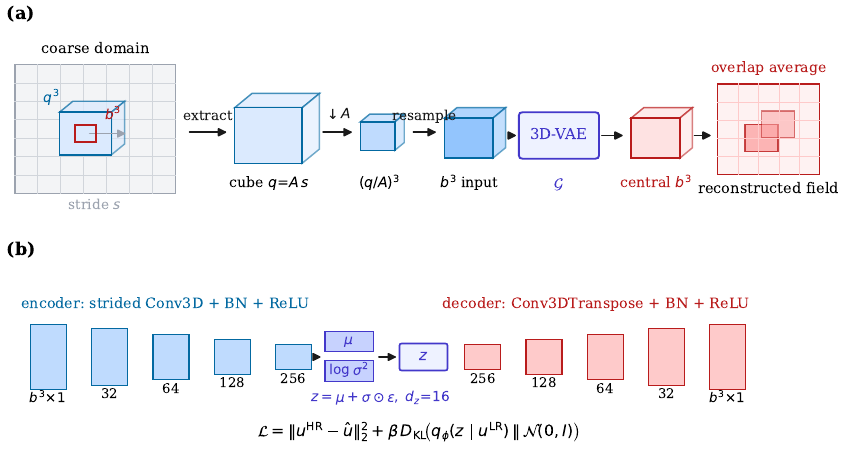}
\caption{Method overview. (a) Patch-based sampling and reconstruction. A
cube of side $q = A s$ is slid over the coarse domain with stride $s$,
downsampled by $A$ and resampled to $b^3 = 16^3$ as network input; the
target is the central $b^3$ sub-block, so the input covers a larger
physical extent than the region reconstructed. Predicted blocks are
reassembled with overlap averaging. Because the learned operator
$\mathcal{G}$ acts on fixed-size blocks, its parameter count is independent
of the domain size. (b) The 3D-VAE: a strided 3D convolutional encoder maps
the input to the parameters of a diagonal Gaussian posterior, a latent
vector of dimension $d_z = 16$ is drawn by the reparameterisation trick,
and a transposed-convolutional decoder reconstructs the $b^3$ block.}
\label{fig:method}
\end{figure*}

\subsection{Problem statement}

Let $u^{\text{HR}}: \Omega \to \mathbb{R}$ denote a DNS-resolution velocity
component on domain $\Omega$, and let $\mathcal{D}_A$ be a downsampling
operator with coarsening factor $A$ in each spatial direction. Given
$u^{\text{LR}} = \mathcal{D}_A u^{\text{HR}}$, we seek an operator
$\mathcal{G}$ with
\[
\mathcal{G}\big(u^{\text{LR}}\big) \approx u^{\text{HR}},
\]
subject to the constraint that the number of parameters of $\mathcal{G}$ is
independent of $|\Omega|$. We satisfy the constraint by learning a
\emph{local} operator on fixed-size blocks and applying it
convolutionally.

\subsection{Patch-based sampling}
\label{sec:sampling}

A cube of side $q$ is slid over the domain with stride $s$ in each of $x$,
$y$, $z$. With coarsening factor $A$ and target block size $b = 16$ we set
\[
q = A \cdot s ,
\]
so that downsampling a $q^3$ cube with stride $A$ yields a low-resolution
input of $(q/A)^3$, which is resampled to $b^3$. The high-resolution target
is the central $b^3$ sub-block of the same cube. Provided $q > b$, the
network therefore sees coarse information from a physically larger
neighbourhood than the region it must reconstruct, which is the operative
design choice: the surrounding context supplies the information needed to
disambiguate small-scale structure, and taking the centre avoids the
boundary effects that would arise if input and target were coextensive. The
procedure and its inverse are given in Algorithm~\ref{alg:patches} and
illustrated in Fig.~\ref{fig:method}(a). Both $A$ and $s$ are
hyperparameters; the results below use the configuration that yields $16^3$
low-resolution inputs, and in the figures the coarse field is displayed on
its own coarser grid, which is why its axis ranges differ from those of the
reference panels.

\begin{algorithm}[t]
\caption{Patch sampling and global reconstruction}
\label{alg:patches}
\begin{algorithmic}[1]
\Require field $u$; coarsening $A$; stride $s$; block size $b = 16$
\State $q \gets A \cdot s$
\State $\mathcal{S} \gets \emptyset$
\For{each corner $(i,j,k)$ with stride $s$ in $x,y,z$}
  \State $C \gets u[i{:}i{+}q,\; j{:}j{+}q,\; k{:}k{+}q]$
  \State $x^{\text{LR}} \gets \textsc{Resample}(C[{::}A,{::}A,{::}A],\, b^3)$
  \State $y^{\text{HR}} \gets \textsc{CentreCrop}(C,\, b^3)$
  \State $\mathcal{S} \gets \mathcal{S} \cup \{(x^{\text{LR}},y^{\text{HR}})\}$
\EndFor
\Statex
\Statex \textbf{Reconstruction}, using the same $A$, $s$, $q$
\State $\hat{u} \gets 0$, \; $w \gets 0$
\For{each corner $(i,j,k)$ with stride $s$}
  \State $p \gets \mathcal{G}\big(\textsc{Resample}(\text{coarse patch},b^3)\big)$
  \State accumulate $p$ into $\hat{u}$ at the centred block; increment $w$
\EndFor
\State \Return $\hat{u} \oslash w$
\end{algorithmic}
\end{algorithm}

\subsection{Network}

The 3D-VAE comprises a strided 3D convolutional encoder, a stochastic
latent bottleneck, and a transposed-convolutional decoder
(Fig.~\ref{fig:method}(b), Table~\ref{tab:arch}). The encoder maps the
$16^3$ input through four convolutional blocks with $32$, $64$, $128$ and
$256$ filters and kernel size $3$, each followed by batch normalisation and
ReLU, to a small spatial extent which is flattened and projected to the
mean $\mu$ and log-variance $\log\sigma^2$ of a diagonal Gaussian
posterior. A latent vector is drawn by the reparameterisation trick
$z = \mu + \sigma \odot \epsilon$, $\epsilon \sim
\mathcal{N}(0,I)$~\cite{kingma2014vae}, and the decoder mirrors the encoder
to return to $16^3$.

\begin{table}[t]
\centering
\caption{3D-VAE architecture. All convolutional and dense layers use batch
normalisation and ReLU except the decoder output.}
\label{tab:arch}
\begin{tabular}{llc}
\toprule
\textbf{Stage} & \textbf{Layer} & \textbf{Filters / units} \\
\midrule
\multirow{7}{*}{Encoder}
 & Input $16^3 \times 1$        & --- \\
 & Conv3D, $k\!=\!3$            & 32 \\
 & Conv3D, $k\!=\!3$, strided   & 64 \\
 & Conv3D, $k\!=\!3$, strided   & 128 \\
 & Conv3D, $k\!=\!3$            & 256 \\
 & Flatten $+$ Dense            & --- \\
 & Dense $\mu$, Dense $\log\sigma^2$ & 16 \\
\midrule
Latent & $z = \mu + \sigma \odot \epsilon$ & $d_z = 16$ \\
\midrule
\multirow{6}{*}{Decoder}
 & Dense $+$ Reshape            & --- \\
 & Conv3DTranspose, strided     & 256 \\
 & Conv3DTranspose, strided     & 128 \\
 & Conv3DTranspose              & 64 \\
 & Conv3DTranspose              & 32 \\
 & Conv3DTranspose (output)     & 1 \\
\bottomrule
\end{tabular}
\end{table}

The latent dimension $d_z = 16$ was selected empirically. Against a
$16^3 = 4096$-dimensional target this is a compression ratio of $256{:}1$,
a strong bottleneck and a plausible mechanism for the small-scale damping
reported in Sec.~\ref{sec:limitations}.

\subsection{Objective and reconstruction}

Training minimises the standard VAE objective, a reconstruction term
regularised by the Kullback--Leibler divergence between the approximate
posterior and a standard normal prior,
\begin{equation}
\mathcal{L} = \big\| u^{\text{HR}} - \hat{u} \big\|_2^2
 \;+\; \beta\, D_{\mathrm{KL}}\!\big(q_\phi(z \mid u^{\text{LR}})
 \,\|\, \mathcal{N}(0,I)\big).
\label{eq:loss}
\end{equation}
Separate models were trained for each velocity component. Training and
validation losses converged smoothly.

At inference the coarse domain is traversed with the same $q$, $s$ and $A$
used in training, each patch is passed through the network, and the
resulting $16^3$ blocks are reassembled with overlapping regions averaged,
which suppresses seams at patch boundaries. Using identical sampling
parameters at training and inference is necessary for the reconstruction to
remain aligned with the reference grid.

\section{Data}

\subsection{DNS: Johns Hopkins Turbulence Database}

We use the JHTDB turbulent channel flow
dataset~\cite{lee2013petascale,graham2016web}, a DNS of incompressible flow
between parallel walls at friction Reynolds number
$Re_\tau \approx 1000$, with no-slip conditions at the walls and
periodicity in the streamwise ($x$) and spanwise ($z$) directions. The
simulation uses a Fourier--Galerkin pseudo-spectral discretisation in the
wall-parallel directions and seventh-order B-spline collocation in the
wall-normal direction, advanced by a low-storage third-order Runge--Kutta
scheme~\cite{lee2015dns}. Parameters are listed in
Table~\ref{tab:dns_params}.

\begin{table}[t]
\centering
\caption{JHTDB turbulent channel flow DNS parameters}
\label{tab:dns_params}
\begin{tabular}{ll}
\toprule
\textbf{Parameter} & \textbf{Value} \\
\midrule
Friction Reynolds number & $Re_\tau \approx 1000$ \\
Domain ($x \times y \times z$) & $8\pi h \times 2h \times 3\pi h$ \\
Grid resolution & $2048 \times 512 \times 1536$ \\
DNS time step & $0.0013$ \\
Stored every & $0.0065$ time units \\
Total time span & $[0,\ 25.9935]$ \\
Training snapshot & frame 3200 \\
Test snapshot & frame 3280 \\
\bottomrule
\end{tabular}
\end{table}

Training samples are drawn from frame 3200 and test samples from frame
3280, both in the statistically stationary regime. The two frames are 80
stored steps apart, i.e. $80 \times 0.0065 = 0.52$ in units of $h/U_b$.
Since $U_b/u_\tau \approx 20$ at $Re_\tau \approx
1000$~\cite{lee2015dns}, this corresponds to approximately $0.026$
eddy-turnover times $h/u_\tau$, or $t^+ \approx 26$ in viscous units.

This separation is short. Near-wall streaks persist for
$t^+ = \mathcal{O}(10^2\!-\!10^3)$, so the test snapshot is not an
independent realisation of the flow: it remains correlated with the
training snapshot, particularly near the wall. We therefore make no claim
of temporal generalisation. The held-out snapshot measures whether the
learned local operator applies to patch configurations absent from the
training set, in the sense of the single-snapshot analysis of Fukami and
Taira~\cite{fukami2024single}, and not whether the model extrapolates in
time. Temporal generalisation is left to future work
(Sec.~\ref{sec:future}).

Results are reported for the streamwise component $u$. Separate models were
trained for $v$ and $w$; their behaviour was qualitatively similar and is
omitted for space.

\subsection{Coarse finite-element simulation}

To test transfer beyond filtered DNS, channel flow was also simulated with
the Oasis solver~\cite{mortensen2015oasis}, a FEniCS-based finite-element
Navier--Stokes code, in the same geometry and with the same boundary
conditions, driven by a constant streamwise pressure gradient
(Table~\ref{tab:les_params}). No explicit subgrid-scale model was
employed: these are under-resolved simulations, that is, implicit LES, and
are labelled ``LES'' in the figures and tables on that understanding. This
is deliberate. Because the finite-element discretisation differs from the
spectral DNS on which the network was trained, these fields provide a
stronger transfer test than filtered DNS, in which input and training
distributions match by construction.

\begin{table}[t]
\centering
\caption{Coarse finite-element (Oasis) channel flow parameters}
\label{tab:les_params}
\begin{tabular}{ll}
\toprule
\textbf{Parameter} & \textbf{Value} \\
\midrule
Domain ($x \times y \times z$) & $8\pi h \times 2h \times 3\pi h$ \\
Coarse grid & $160 \times 120 \times 40$ \\
Fine grid & $320 \times 240 \times 80$ \\
Time step & $0.005$ \\
Output frequency & every $0.1$ time units \\
Total time span & $[0,\ 25]$ \\
SGS model & none (implicit LES) \\
\bottomrule
\end{tabular}
\end{table}

Snapshots were extracted at $20$ time units for training and $21$ time
units for testing, avoiding temporal overlap between the two sets.

\section{Results}
\label{sec:results}

\subsection{Patch-level reconstruction}

\begin{figure}[t]
\centering
\includegraphics[width=\linewidth]{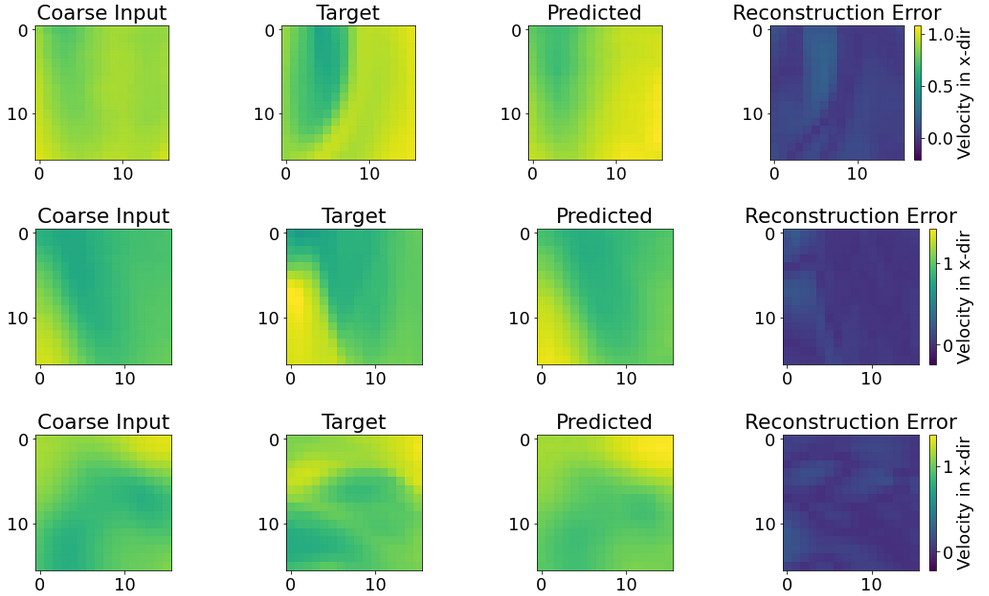}
\caption{Conditional 3D-GAN. Coarse input, target, prediction, and
pointwise reconstruction error on randomly selected orthogonal planes
through three $16^3$ test patches, streamwise velocity.}
\label{fig:gan_patches}
\end{figure}

\begin{figure}[t]
\centering
\includegraphics[width=\linewidth]{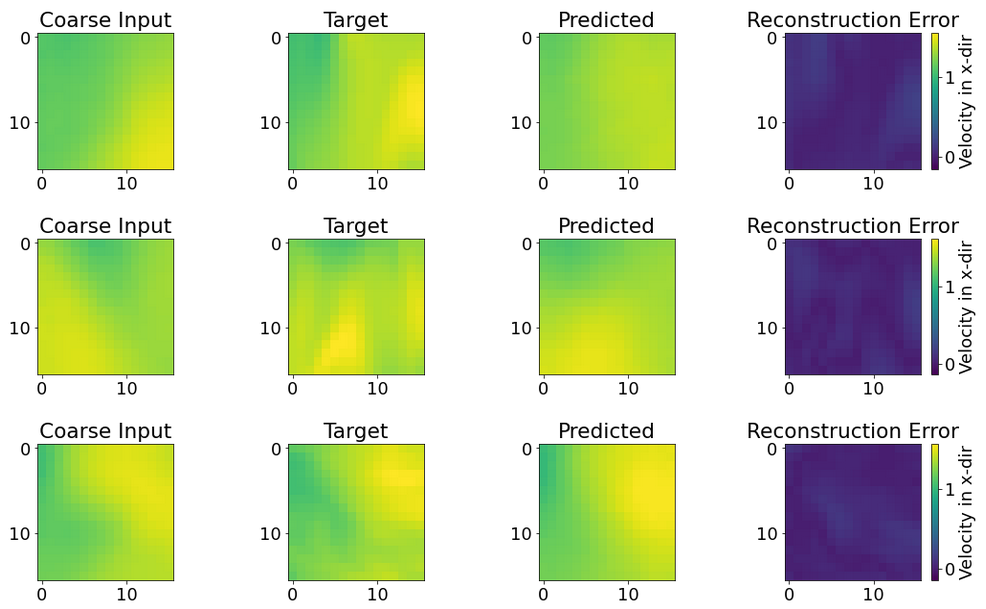}
\caption{3D-VAE, plotted as in Fig.~\ref{fig:gan_patches}.}
\label{fig:vae_patches}
\end{figure}

Figs.~\ref{fig:gan_patches} and~\ref{fig:vae_patches} show coarse input,
target, prediction and pointwise reconstruction error for randomly selected
orthogonal planes through $16^3$ test patches, for the conditional 3D-GAN
and the 3D-VAE respectively. The 3D-VAE recovers the large-scale structure
of the target and the bulk of its intermediate-scale variation; the 3D-GAN
does not, and is discussed separately in Sec.~\ref{sec:gan}.

\subsection{Reconstruction of the global field from filtered DNS}

\begin{figure}[t]
\centering
\includegraphics[width=\linewidth]{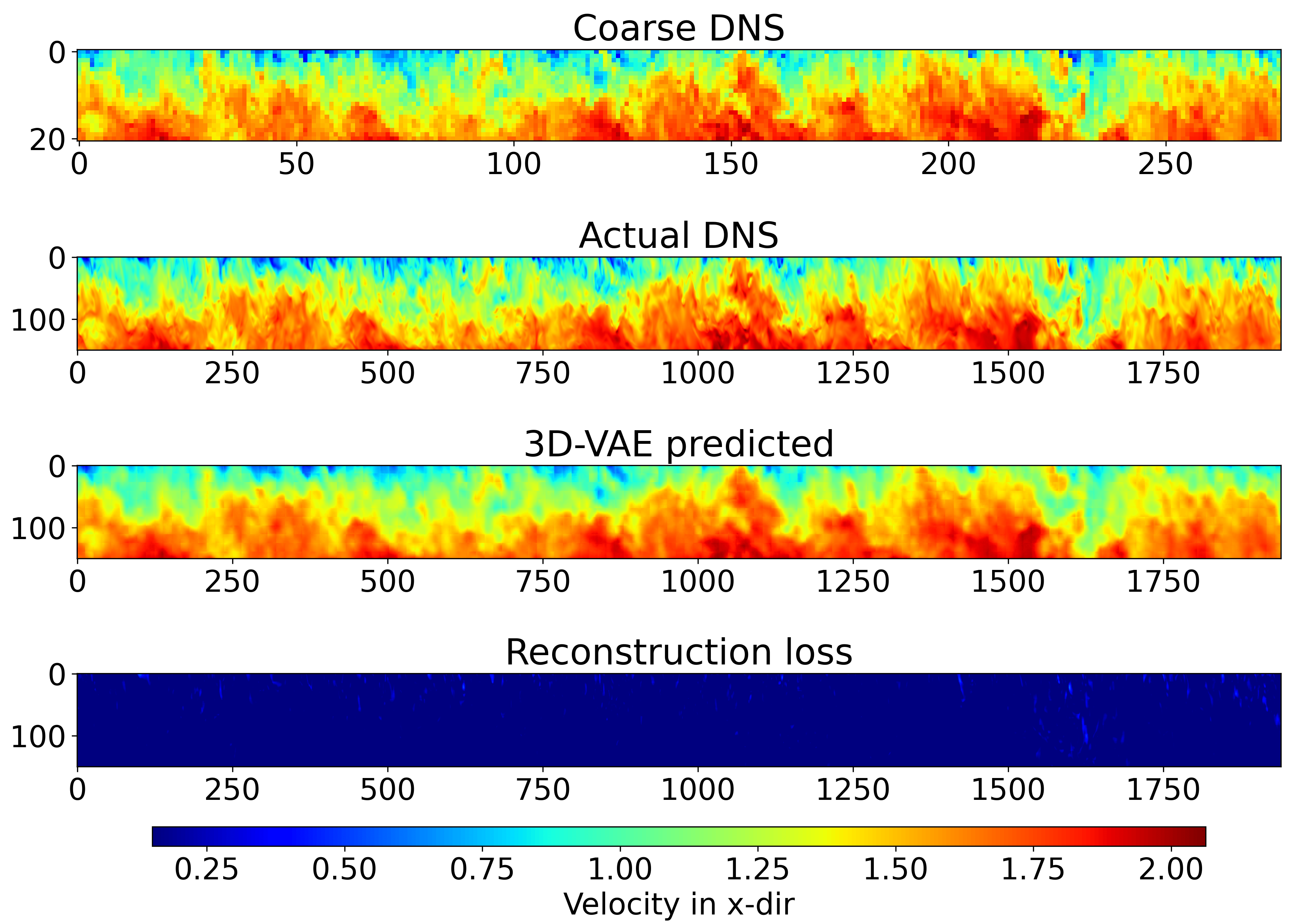}
\caption{Mid-channel $x$--$y$ plane, streamwise velocity: coarse DNS input,
DNS reference, 3D-VAE reconstruction, and pointwise reconstruction error.
The coarse panel is displayed on its own coarser grid.}
\label{fig:xy_real}
\end{figure}

Patch predictions are reassembled into the full channel domain as described
in Sec.~\ref{sec:sampling}. Fig.~\ref{fig:xy_real} compares, on the
mid-channel $x$--$y$ plane, the coarse input, the DNS reference, the
3D-VAE reconstruction and the pointwise error;
Table~\ref{tab:amplitude_range} lists the corresponding value ranges.

\begin{table}[t]
\centering
\caption{Streamwise velocity range on the mid-channel $x$--$y$ plane,
filtered DNS input. $\varepsilon_{\max}$ is the maximum pointwise error
against the DNS reference over the plane.}
\label{tab:amplitude_range}
\begin{tabular}{lcc}
\toprule
\textbf{Field} & \textbf{Min} & \textbf{Max} \\
\midrule
Coarse DNS input & 0.2995 & 1.9982 \\
DNS reference & 0.1511 & 2.0627 \\
3D-VAE & 0.4242 & 2.0097 \\
\midrule
\multicolumn{1}{l}{$\varepsilon_{\max}$, 3D-VAE} & \multicolumn{2}{c}{0.6001} \\
\bottomrule
\end{tabular}
\end{table}

The 3D-VAE recovers the DNS maximum to within $2.6\%$ but does not
reproduce the extreme minimum, consistent with a model that
under-predicts the tails of the velocity distribution.

\subsection{Comparison against interpolation}

\begin{figure}[t]
\centering
\includegraphics[width=\linewidth]{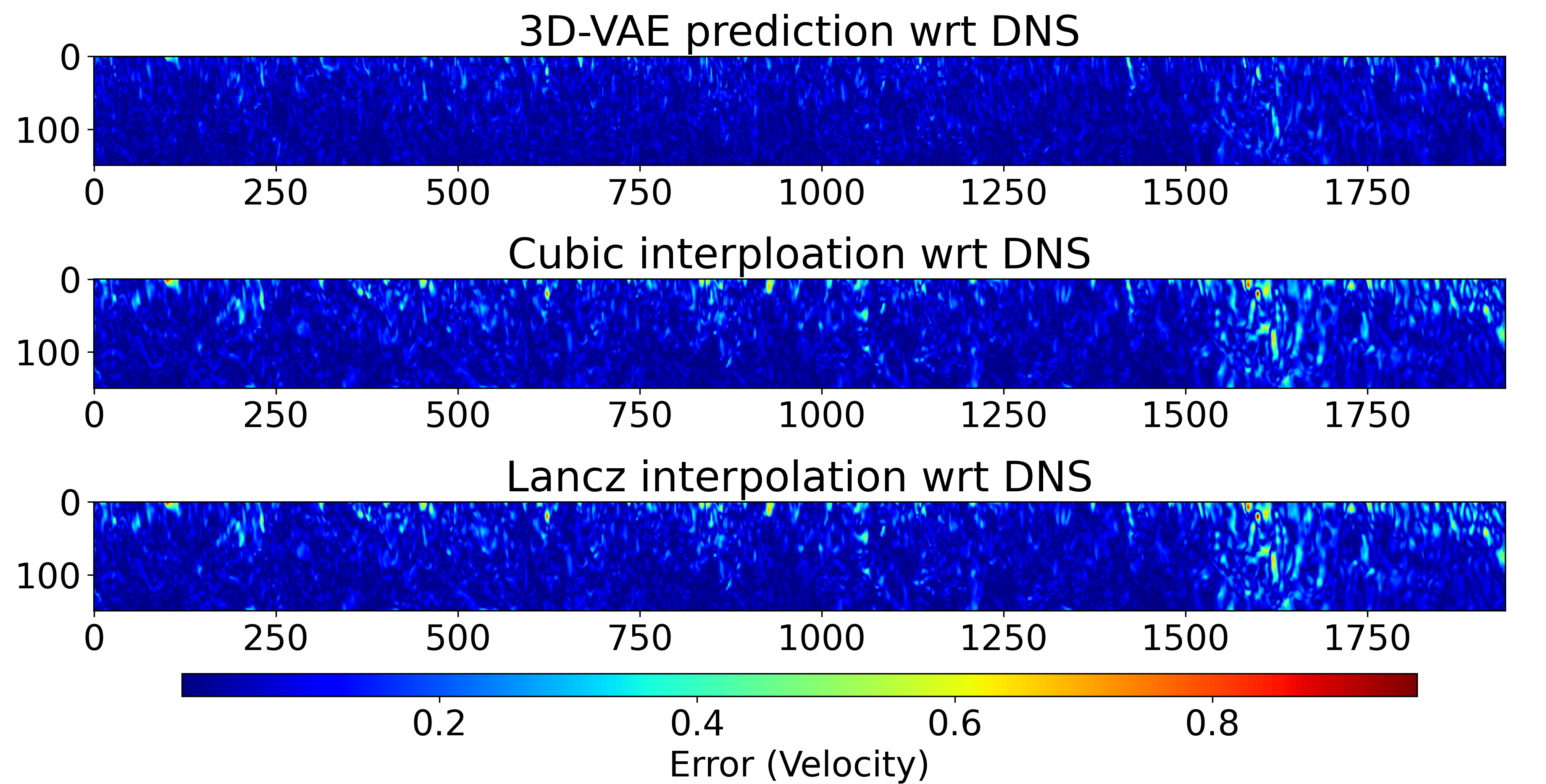}
\caption{Pointwise error against DNS for the 3D-VAE, tricubic and Lanczos
interpolation, mid-channel $x$--$y$ plane.}
\label{fig:interp_comparison}
\end{figure}

Table~\ref{tab:err_physical} and Fig.~\ref{fig:interp_comparison} compare
the 3D-VAE against tricubic and Lanczos interpolation applied to the same
coarse input. The 3D-VAE reduces mean absolute error by $27\%$ relative to
tricubic interpolation and maximum absolute error by $36\%$.

\begin{table}[t]
\centering
\caption{Physical-space error against DNS, streamwise velocity,
mid-channel $x$--$y$ plane}
\label{tab:err_physical}
\begin{tabular}{lcc}
\toprule
\textbf{Method} & \textbf{Max $|$error$|$} & \textbf{Mean $|$error$|$} \\
\midrule
3D-VAE  & \textbf{0.6001} & \textbf{0.0547} \\
Tricubic & 0.9433 & 0.0749 \\
Lanczos & 0.9589 & 0.0759 \\
\bottomrule
\end{tabular}
\end{table}

\subsection{Spectral analysis}

\begin{figure}[t]
\centering
\includegraphics[width=\linewidth]{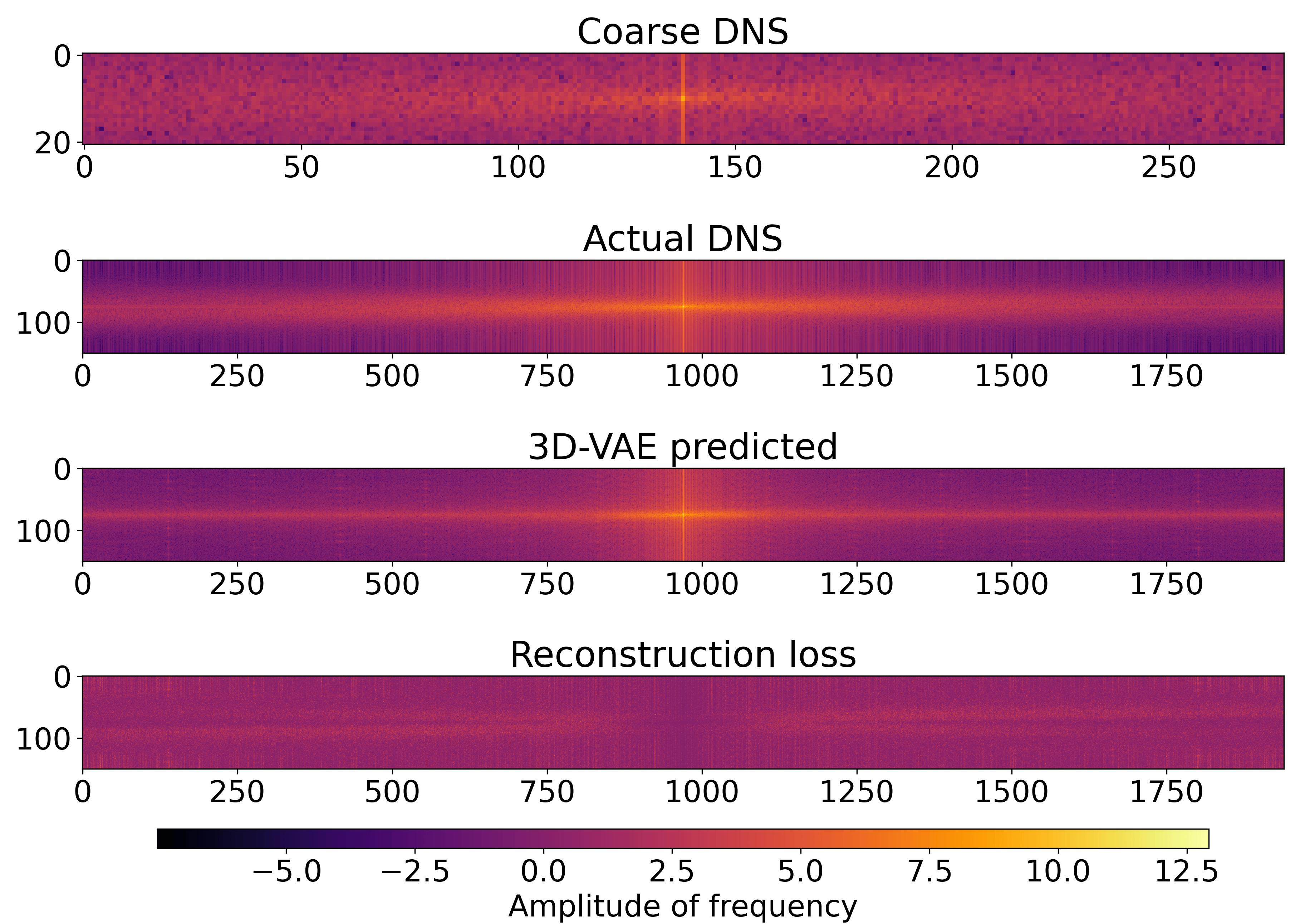}
\caption{FFT amplitude, $\ln|\hat{u}|$, on the mid-channel $x$--$y$ plane
for the coarse DNS input, DNS reference, 3D-VAE prediction, and the
amplitude of the reconstruction error.}
\label{fig:fft_amplitude}
\end{figure}

\begin{figure}[t]
\centering
\includegraphics[width=\linewidth]{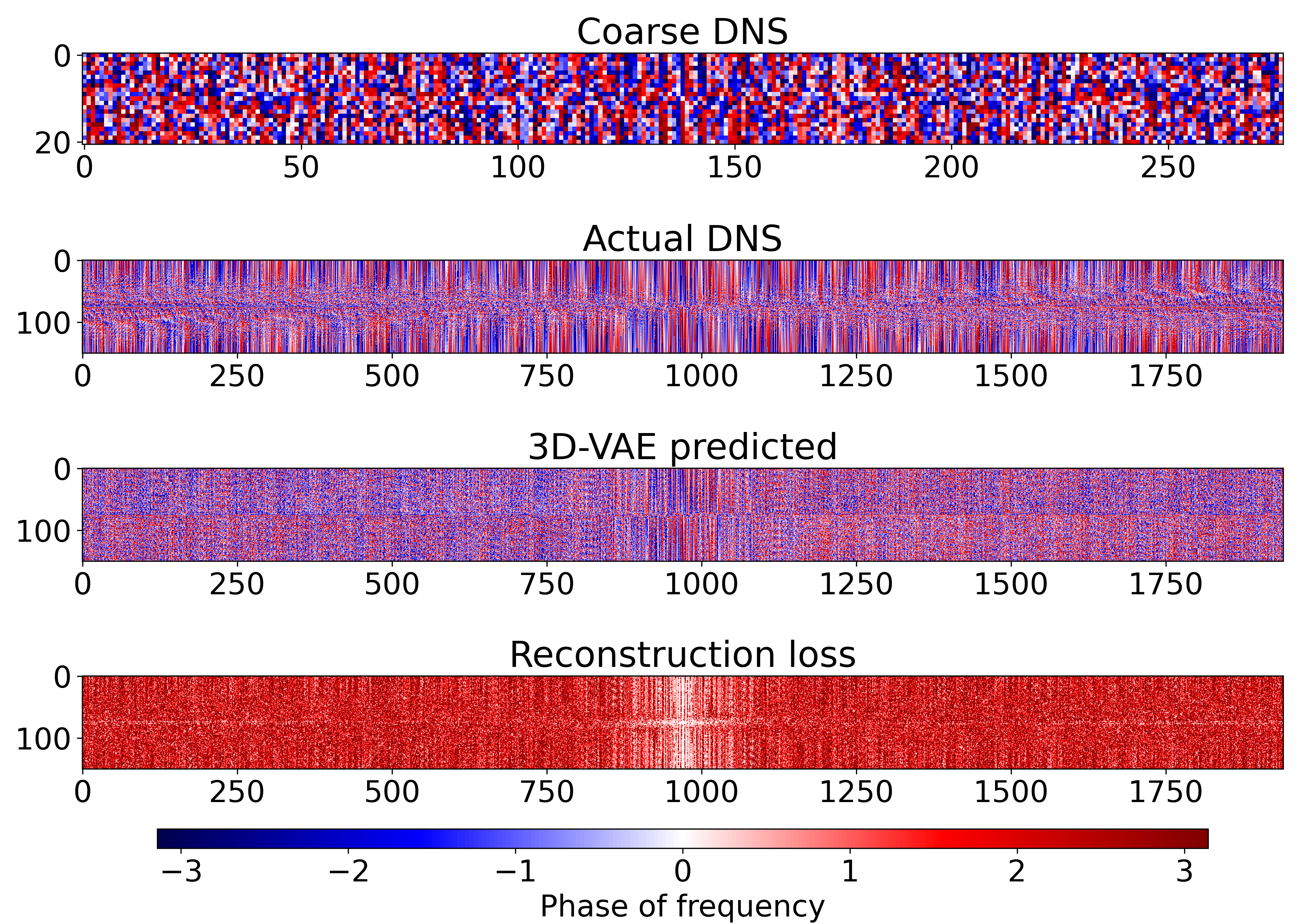}
\caption{FFT phase on the mid-channel $x$--$y$ plane, plotted as in
Fig.~\ref{fig:fft_amplitude}. Periodic structure at the patch stride is
visible in the 3D-VAE panel.}
\label{fig:fft_phase}
\end{figure}

Pointwise error is a weak diagnostic for turbulence: two fields with
similar $L_2$ error can distribute energy across scales very differently,
and it is the distribution that determines whether a reconstruction is
physically usable. We therefore compute the two-dimensional FFT of the
mid-channel $x$--$y$ plane for the coarse input, the DNS reference and the
3D-VAE output, and compare amplitude
(Fig.~\ref{fig:fft_amplitude}, Table~\ref{tab:fft_amplitude}) and phase
(Fig.~\ref{fig:fft_phase}) separately. Amplitude governs the energy
cascade; phase governs the spatial placement of vortices, streaks and
shear layers, and a reconstruction can match the amplitude spectrum while
misplacing every structure in it. Amplitudes are reported as
$\ln|\hat{u}|$ of the unnormalised transform, so the negative values in
Table~\ref{tab:fft_amplitude} are logarithms rather than amplitudes.

\begin{table}[t]
\centering
\caption{FFT amplitude $\ln|\hat{u}|$ on the mid-channel $x$--$y$ plane}
\label{tab:fft_amplitude}
\begin{tabular}{lcc}
\toprule
\textbf{Field} & \textbf{Min} & \textbf{Max} \\
\midrule
Coarse DNS input & $-3.41$ & $8.99$ \\
DNS reference & $-7.50$ & $12.91$ \\
3D-VAE & $-5.40$ & $12.92$ \\
\midrule
\multicolumn{1}{l}{$\varepsilon_{\max}$, 3D-VAE} & \multicolumn{2}{c}{$7.27$} \\
\bottomrule
\end{tabular}
\end{table}

\begin{table}[t]
\centering
\caption{Spectral amplitude error against DNS, mid-channel $x$--$y$ plane}
\label{tab:err_spectral}
\begin{tabular}{lcc}
\toprule
\textbf{Method} & \textbf{Max $|$error$|$} & \textbf{Mean $|$error$|$} \\
\midrule
3D-VAE  & \textbf{7.2714}  & \textbf{0.9062} \\
Tricubic & 10.9714 & 2.6259 \\
Lanczos & 9.0570  & 2.8463 \\
\bottomrule
\end{tabular}
\end{table}

\begin{figure}[t]
\centering
\includegraphics[width=\linewidth]{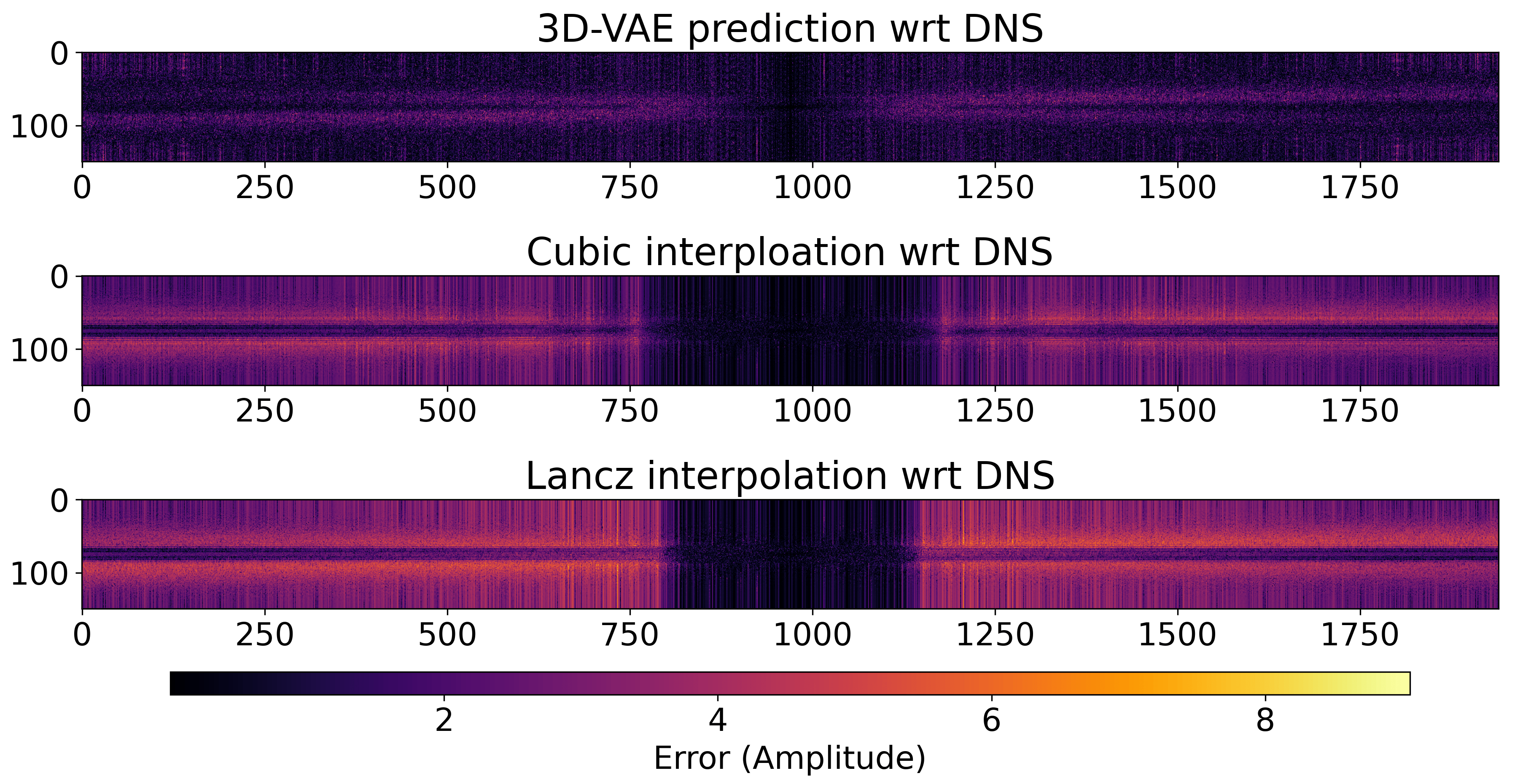}
\caption{Spectral amplitude error against DNS for the 3D-VAE, tricubic and
Lanczos interpolation, mid-channel $x$--$y$ plane.}
\label{fig:vae_vs_interp_fft}
\end{figure}

\begin{figure*}[t]
\centering
\includegraphics[width=\linewidth]{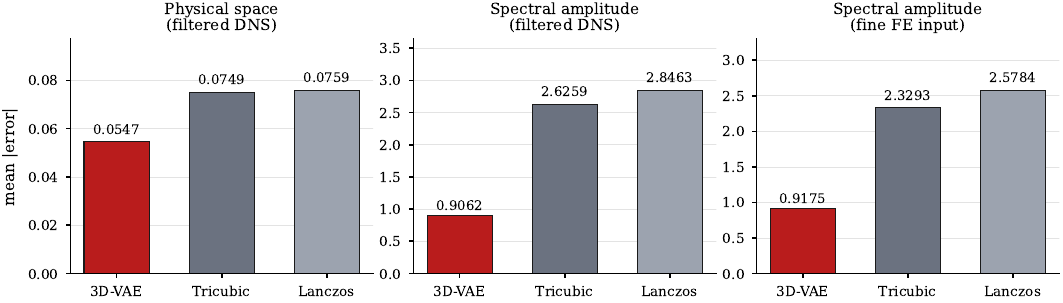}
\caption{Mean absolute error against DNS for the 3D-VAE and the two
interpolation baselines, in physical space and in spectral amplitude, for
both the filtered-DNS and the coarse finite-element input. The advantage
over interpolation is consistently larger in spectral space than in
physical space. Values are those tabulated in
Tables~\ref{tab:err_physical}, \ref{tab:err_spectral} and
\ref{tab:fine_err}.}
\label{fig:error_summary}
\end{figure*}

The gap between methods is far larger in spectral space than in physical
space. Fig.~\ref{fig:error_summary} collects the comparison: the 3D-VAE
reduces mean spectral amplitude error by roughly a factor of three, against
a $27\%$ reduction in pointwise error. This is the central quantitative
result, and the mechanism is straightforward. Interpolation cannot create
wavenumber content above the coarse Nyquist limit; it can only redistribute
what is already present. The 3D-VAE, having learned the statistics of the
flow, generates content there.

Two caveats follow from the figures: deviation from DNS grows at the
highest wavenumbers, and periodic artefacts appear in the phase plots at
the patch stride. Both are discussed in Sec.~\ref{sec:limitations}.

\IfFileExists{fig_energy_spectrum.pdf}{%
\subsection{Energy spectrum}
\label{sec:spectrum}

The amplitude maps of Fig.~\ref{fig:fft_amplitude} show the full
two-dimensional spectrum but make it difficult to read off how error is
distributed with scale. Fig.~\ref{fig:energy_spectrum} therefore reduces
the same data to the radially averaged spectrum $E(k)$ of the mid-channel
plane for the coarse input, the DNS reference and the 3D-VAE
reconstruction, together with the $k^{-5/3}$ inertial-range slope and the
Nyquist wavenumber of the coarse input.

\begin{figure}[t]
\centering
\includegraphics[width=0.92\linewidth]{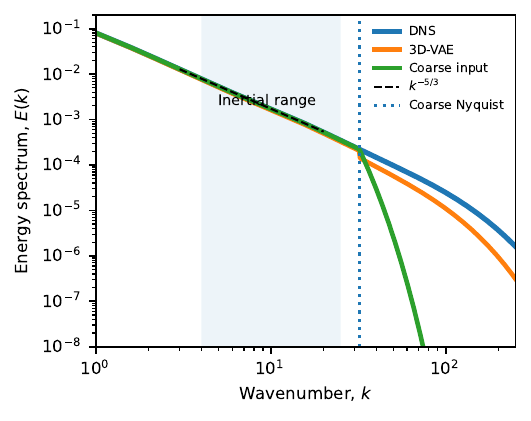}
\caption{Radially averaged energy spectrum $E(k)$ of the mid-channel
$x$--$y$ plane for the coarse input, the DNS reference and the 3D-VAE
reconstruction. The dashed line marks the $k^{-5/3}$ inertial-range slope;
the dotted vertical line marks the Nyquist wavenumber of the coarse input,
above which the input carries no information and all recovered content is
synthesised by the model.}
\label{fig:energy_spectrum}
\end{figure}

This localises in wavenumber the aggregate advantage reported in
Table~\ref{tab:err_spectral}. Below the coarse Nyquist limit the three
fields carry comparable energy: the information is present in the input,
and interpolation is largely sufficient to recover it. Above that limit the
coarse field has no content by construction, so the separation between the
curves in this band is attributable entirely to what the model
synthesises---which is why the spectral error gap
(Fig.~\ref{fig:error_summary}) is so much wider than the pointwise one. The
3D-VAE recovers a substantial fraction of the DNS energy above the coarse
Nyquist limit while falling below the reference at the highest wavenumbers,
which is the spectral signature of the small-scale damping discussed in
Sec.~\ref{sec:limitations}.
}{}

\subsection{Conditional 3D-GAN: a negative result}
\label{sec:gan}

A conditional 3D-GAN was trained on identical data. The generator follows a
U-Net topology of 3D convolution, batch normalisation and LeakyReLU blocks
with skip connections to a transposed-convolution decoder; the critic is a
3D convolutional network returning a scalar score. Training used the
Wasserstein objective with weight clipping~\cite{arjovsky2017wgan}.

The model did not converge to a useful reconstruction on either the
training or the test set (Fig.~\ref{fig:gan_patches}). We report this
rather than omit it. Two candidate explanations are consistent with the
literature. First, the adversarial term trades pointwise accuracy for
high-wavenumber content, and at this patch size and dataset scale that
trade may be unfavourable~\cite{macart2024adversarial}. Second, the models
that do succeed in this setting generally add explicit physical constraints
to the generator loss---continuity and pressure-Poisson residuals in
TEGAN~\cite{subramaniamturbulence}---which we did not. We did not attempt
WGAN with a gradient penalty~\cite{gulrajani2017improved}, which is the
standard remedy for the instability observed and the obvious next step.

\section{Transfer to Coarse Finite-Element Fields}
\label{sec:les2dns}

\begin{figure}[t]
\centering
\begin{subfigure}[b]{0.48\linewidth}
  \centering
  \includegraphics[width=\linewidth]{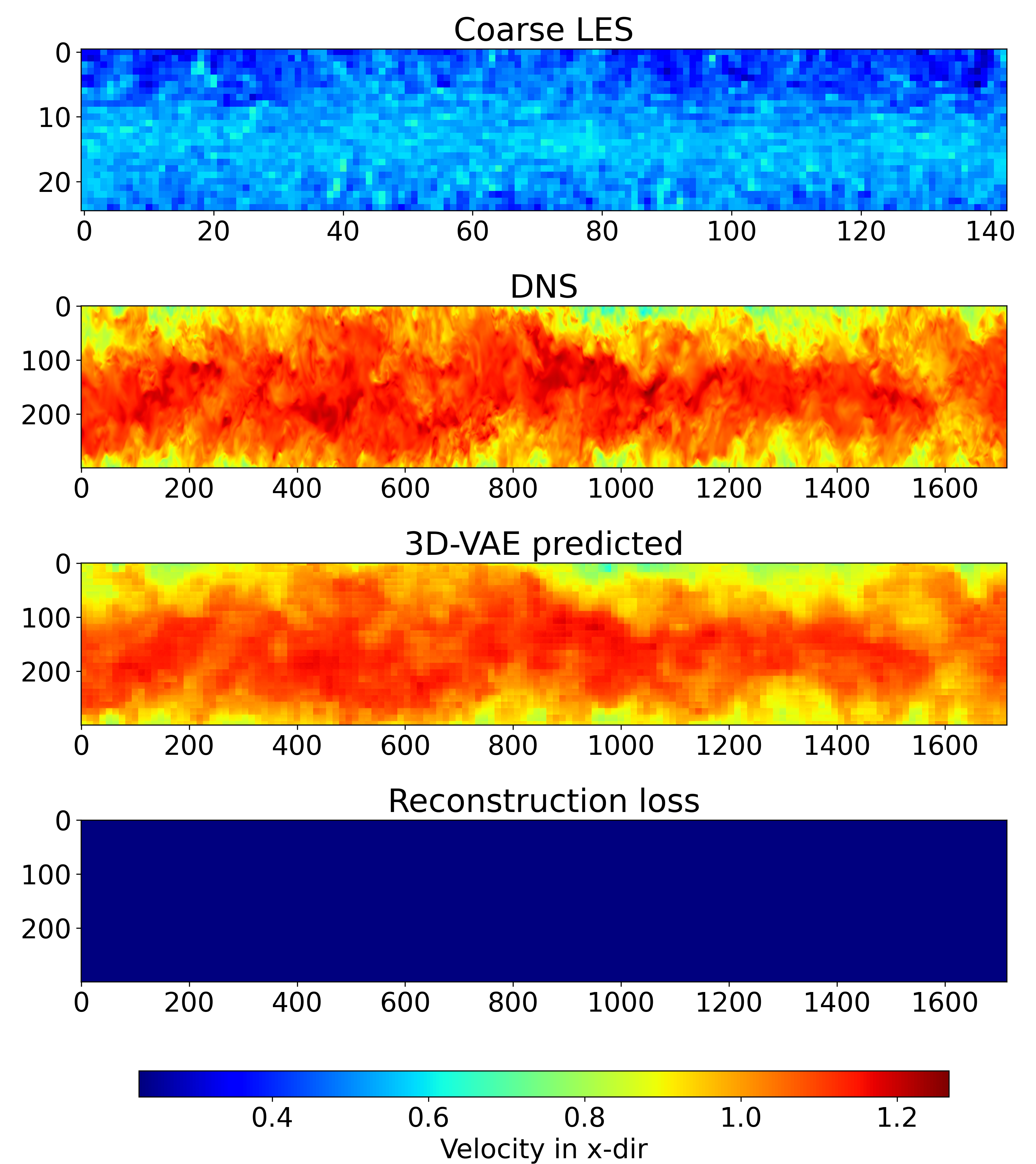}
  \caption{Coarse LES, velocity}
\end{subfigure}\hfill
\begin{subfigure}[b]{0.48\linewidth}
  \centering
  \includegraphics[width=\linewidth]{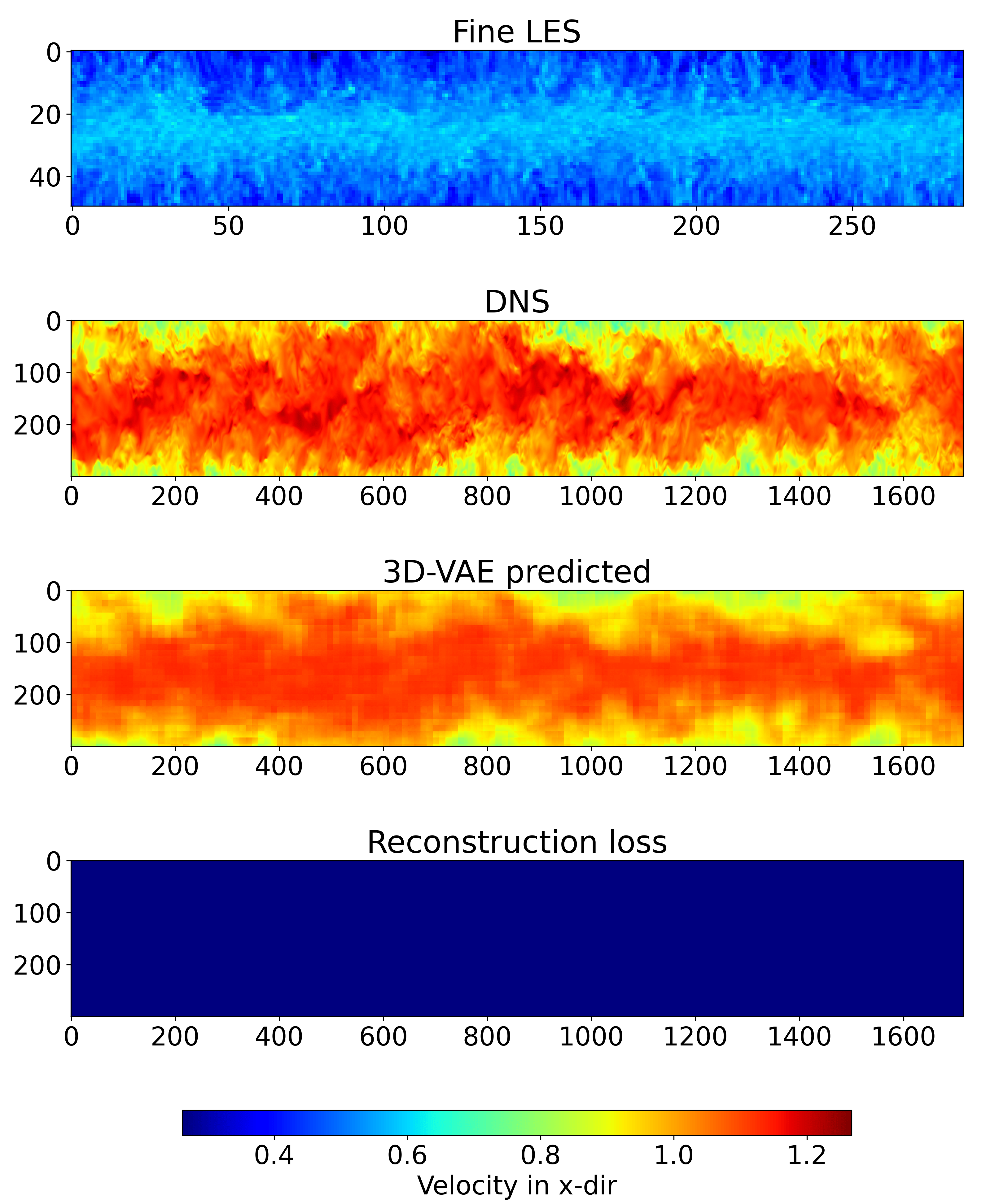}
  \caption{Fine LES, velocity}
\end{subfigure}
\vspace{0.4em}
\begin{subfigure}[b]{0.48\linewidth}
  \centering
  \includegraphics[width=\linewidth]{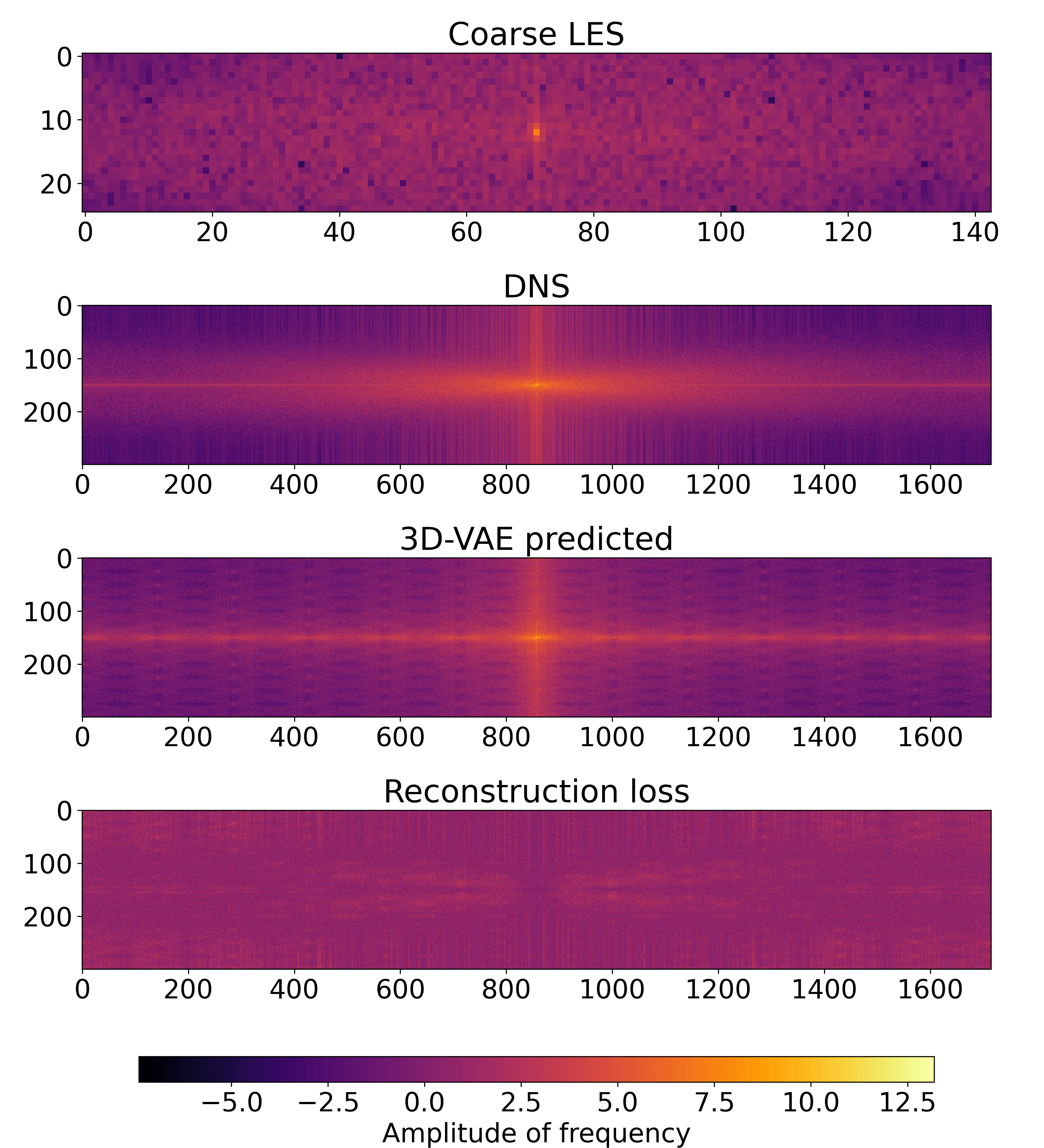}
  \caption{Coarse LES, amplitude}
\end{subfigure}\hfill
\begin{subfigure}[b]{0.48\linewidth}
  \centering
  \includegraphics[width=\linewidth]{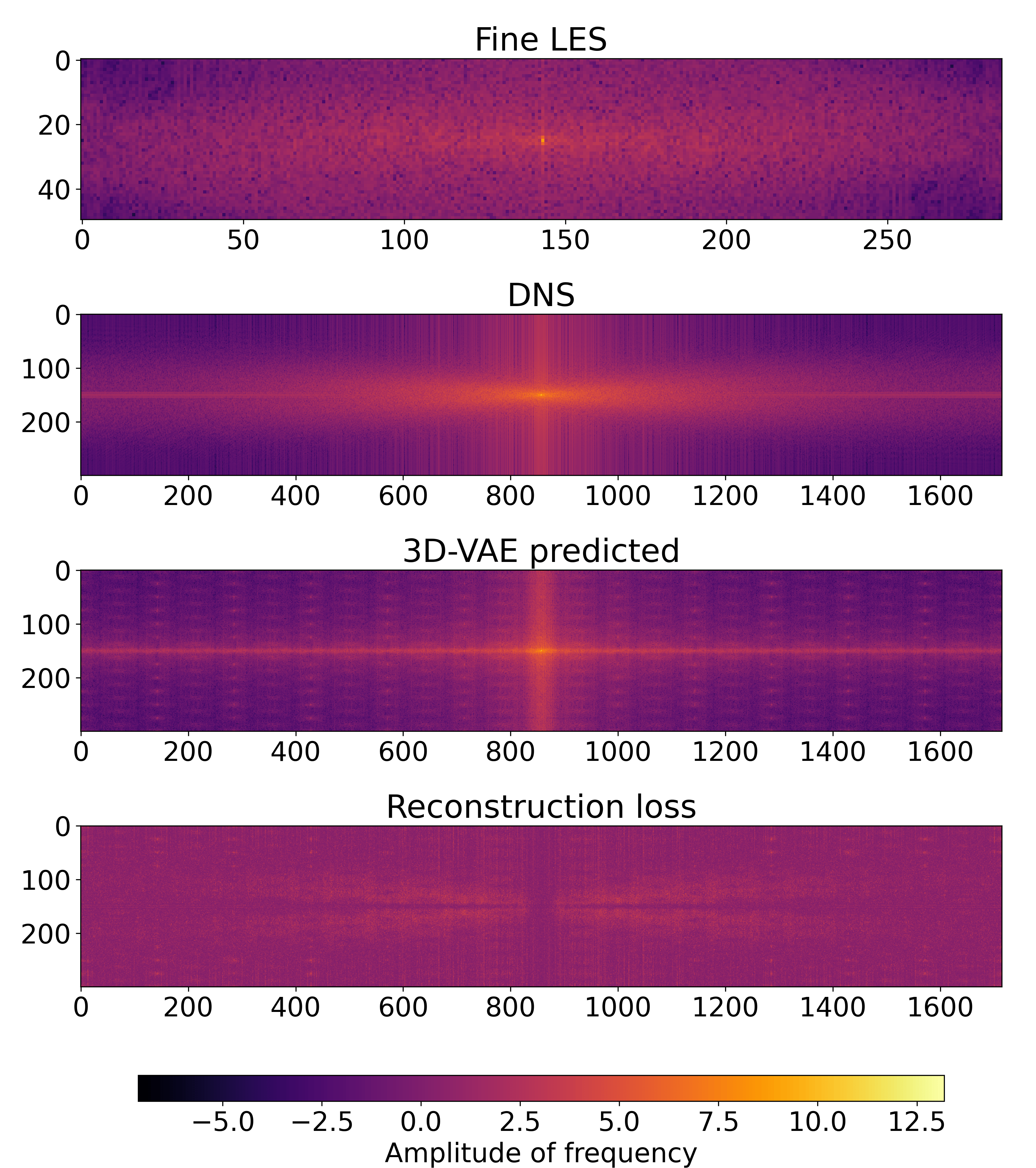}
  \caption{Fine LES, amplitude}
\end{subfigure}
\vspace{0.4em}
\begin{subfigure}[b]{0.48\linewidth}
  \centering
  \includegraphics[width=\linewidth]{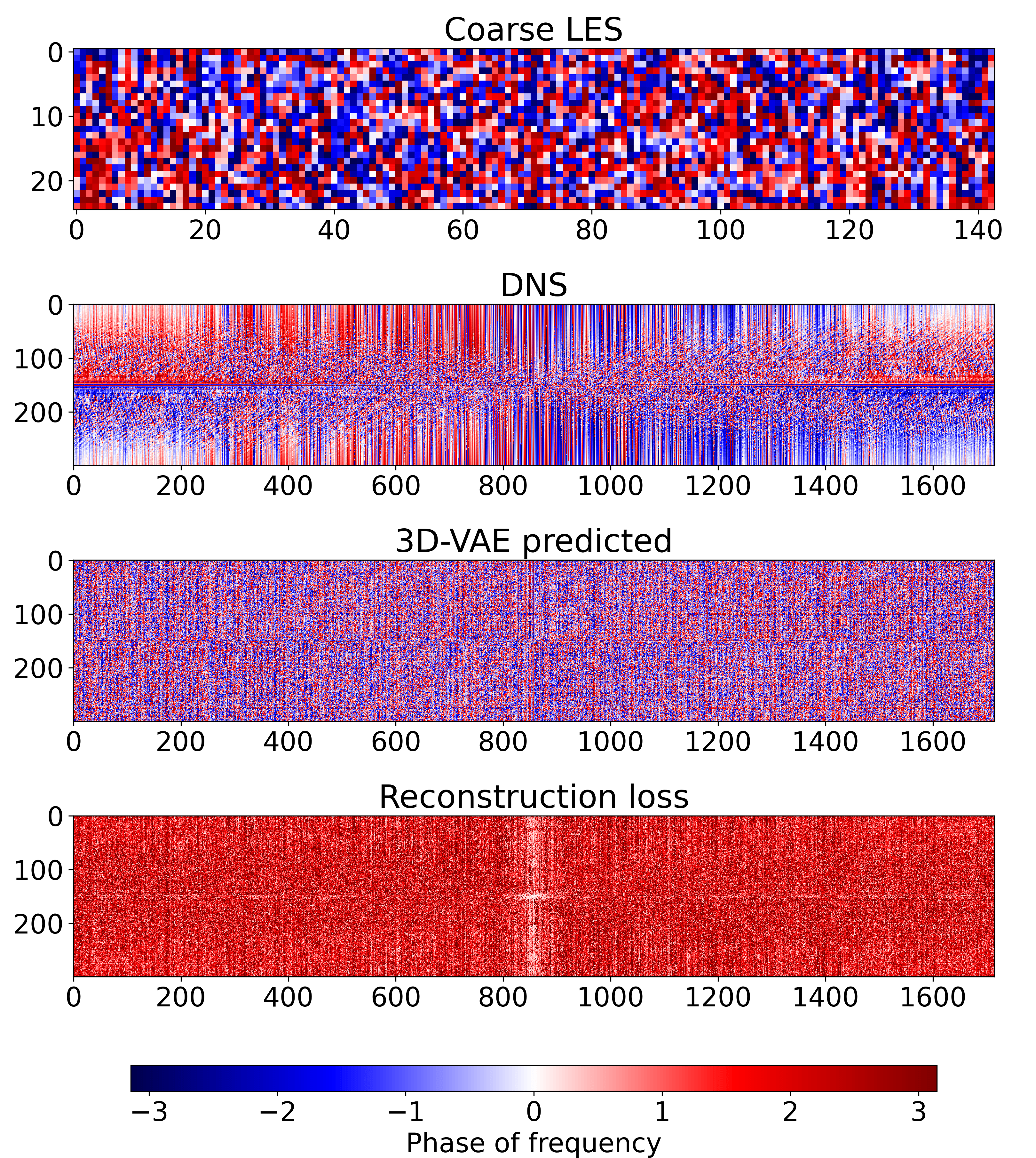}
  \caption{Coarse LES, phase}
\end{subfigure}\hfill
\begin{subfigure}[b]{0.48\linewidth}
  \centering
  \includegraphics[width=\linewidth]{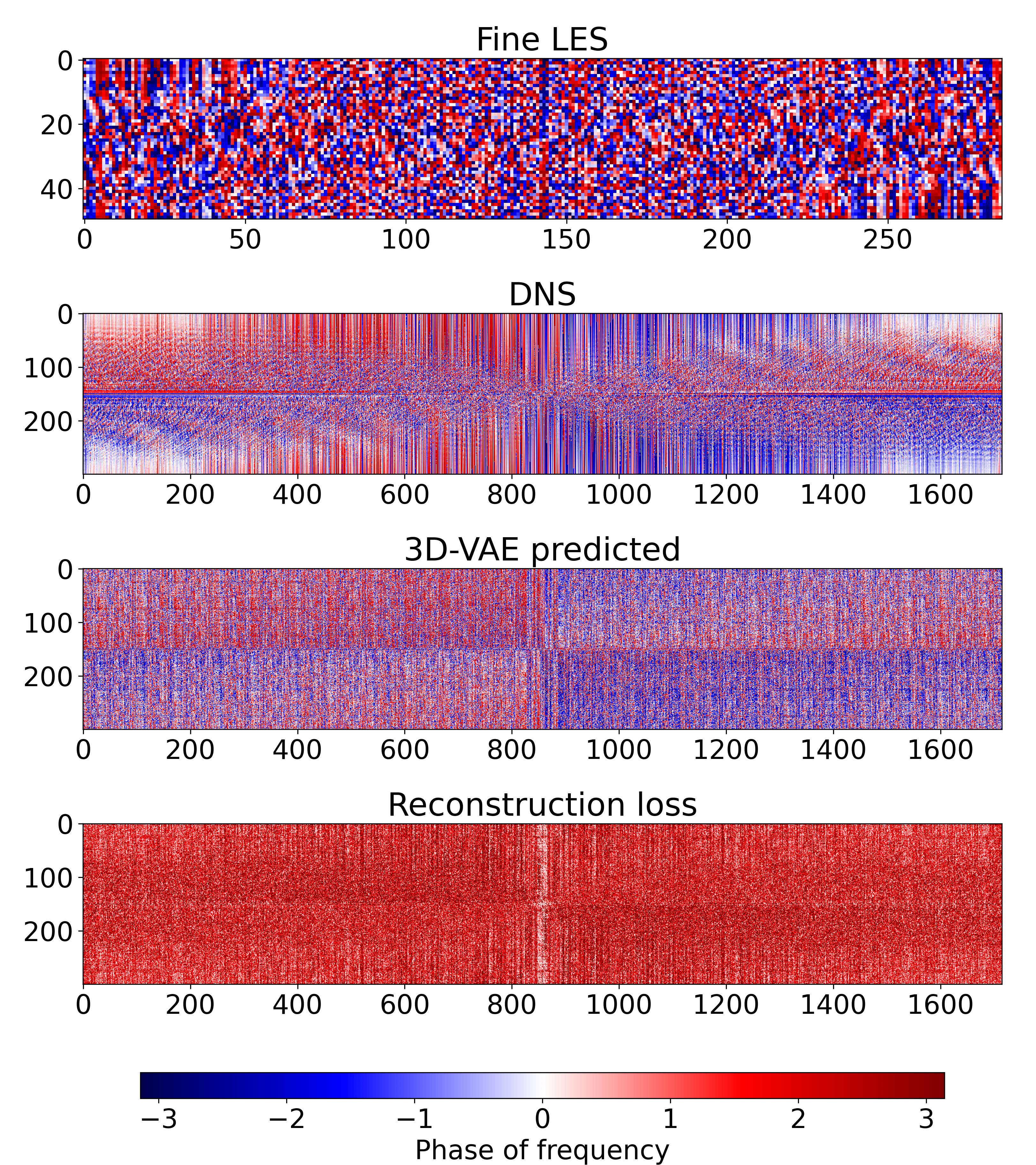}
  \caption{Fine LES, phase}
\end{subfigure}
\caption{Transfer to coarse finite-element input, mid-channel $x$--$y$
plane. Each panel shows, from top to bottom, the finite-element input, the
DNS reference, the 3D-VAE reconstruction, and the reconstruction error.
Input panels are displayed on their own coarser grids.}
\label{fig:fe_transfer}
\end{figure}

The results above super-resolve fields obtained by filtering DNS with the
same operator used to build the training set. A more demanding test is
whether the model transfers to coarse fields produced by a different
numerical method. We apply the DNS-trained 3D-VAE to the Oasis fields of
Table~\ref{tab:les_params}, on snapshots temporally separated from those
used in training. Table~\ref{tab:fe_ranges} reports value ranges in
physical space and in spectral amplitude for both the coarse and fine
finite-element grids, and Fig.~\ref{fig:fe_transfer} shows the
corresponding fields and spectra.

\begin{table}[t]
\centering
\caption{Transfer to coarse finite-element input: value ranges on the
mid-channel $x$--$y$ plane. $\varepsilon_{\max}$ is the maximum pointwise
error against the DNS reference over the plane.}
\label{tab:fe_ranges}
\begin{tabular}{llccc}
\toprule
\textbf{Case} & \textbf{Field} & \textbf{Min} & \textbf{Max} &
$\boldsymbol{\varepsilon_{\max}}$ \\
\midrule
\multicolumn{5}{l}{\emph{Physical space}} \\
\multirow{3}{*}{Coarse LES} & Input    & 0.2290 & 0.6745 & \\
                            & DNS ref. & 0.6020 & 1.2664 & \\
                            & 3D-VAE   & 0.6118 & 1.1929 & 0.2118 \\
\cmidrule(l){2-5}
\multirow{3}{*}{Fine LES}   & Input    & 0.2622 & 0.6745 & \\
                            & DNS ref. & 0.6349 & 1.2669 & \\
                            & 3D-VAE   & 0.7536 & 1.1398 & 0.2192 \\
\midrule
\multicolumn{5}{l}{\emph{FFT amplitude, $\ln|\hat{u}|$}} \\
\multirow{3}{*}{Coarse LES} & Input    & $-4.5221$ & $7.4949$ & \\
                            & DNS ref. & $-7.3919$ & $13.1899$ & \\
                            & 3D-VAE   & $-6.8298$ & $13.1755$ & $7.9236$ \\
\cmidrule(l){2-5}
\multirow{3}{*}{Fine LES}   & Input    & $-4.7658$ & $8.9022$ & \\
                            & DNS ref. & $-7.1272$ & $13.1889$ & \\
                            & 3D-VAE   & $-7.0866$ & $13.1835$ & $7.2220$ \\
\bottomrule
\end{tabular}
\end{table}

\begin{table}[t]
\centering
\caption{Spectral amplitude error against DNS, fine finite-element input}
\label{tab:fine_err}
\begin{tabular}{lcc}
\toprule
\textbf{Method} & \textbf{Max $|$error$|$} & \textbf{Mean $|$error$|$} \\
\midrule
3D-VAE  & \textbf{7.2220} & \textbf{0.9175} \\
Tricubic & 9.0866 & 2.3293 \\
Lanczos & 9.1416 & 2.5784 \\
\bottomrule
\end{tabular}
\end{table}

\begin{figure}[t]
\centering
\includegraphics[width=\linewidth]{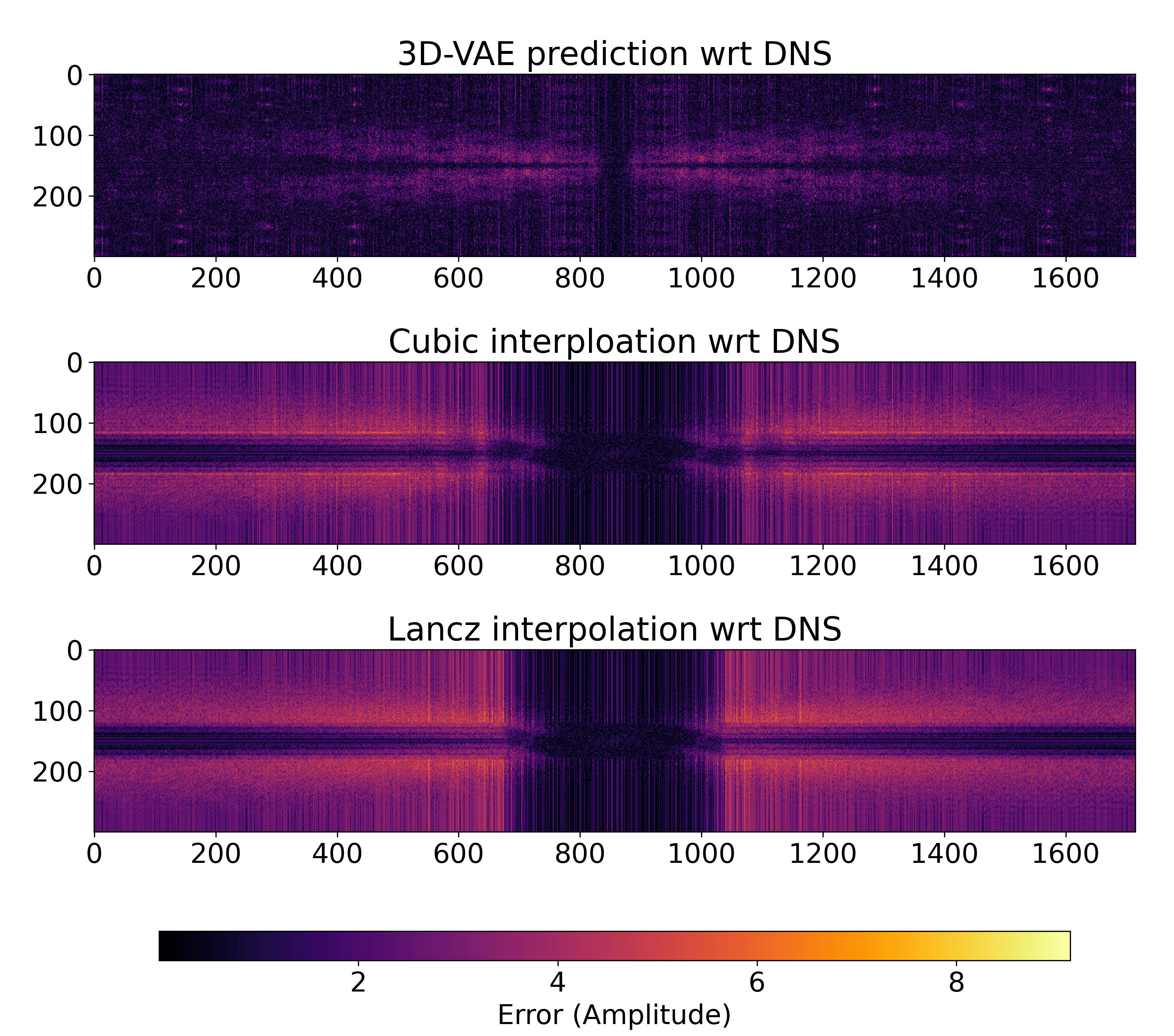}
\caption{Spectral amplitude error against DNS for the 3D-VAE, tricubic and
Lanczos interpolation, fine finite-element input.}
\label{fig:xy_fine_comp_fft}
\end{figure}

The 3D-VAE recovers spectral content that is absent from the
finite-element input but present in the DNS reference, and the advantage
over interpolation persists at approximately the same ratio as in the
filtered-DNS case ($0.92$ against $2.33$ and $2.58$ mean spectral error,
Table~\ref{tab:fine_err}).

One feature of Table~\ref{tab:fe_ranges} requires comment. The
finite-element fields span roughly half the velocity range of the DNS
reference: the coarse input covers $[0.229, 0.675]$ against
$[0.602, 1.266]$ for DNS, a factor of about two in the midpoint of the
range. The two datasets are therefore not at the same bulk velocity or
normalisation. Part of what the network performs in this transfer setting
is consequently an amplitude correction in addition to spectral
enrichment, and the physical-space errors in this section should be read
with that in mind. The spectral comparison, which concerns the
\emph{distribution} of energy across wavenumbers rather than its absolute
level, is the more informative half of this experiment.

\section{Limitations}
\label{sec:limitations}

\subsection{Failure modes}

\textbf{Damping of the smallest scales.} Deviation from DNS grows with
wavenumber (Fig.~\ref{fig:fft_amplitude}). Two mechanisms plausibly
contribute: the coarse input carries no information above its own Nyquist
limit, so the highest wavenumbers must be inferred from learned statistics
rather than reconstructed; and the $256{:}1$ latent bottleneck penalises
exactly the low-variance, high-wavenumber components. The reconstruction
should therefore be understood as statistically plausible at small scales
rather than pointwise correct there.

\textbf{Periodic artefacts.} The phase spectra
(Figs.~\ref{fig:fft_phase} and~\ref{fig:fe_transfer}) show repeating
structure at the patch stride, an artefact of assembling the field from
fixed-size $16^3$ blocks. Overlap averaging reduces but does not eliminate
it. Larger patches would capture longer-range correlations and weaken the
artefact, at a cost in memory and in the volume of training data required.
This is the accuracy--feasibility trade-off that motivated the patch-based
design in the first place, and we have not measured where its optimum
lies.

\textbf{Tails of the distribution.} The model systematically fails to
reach the DNS minimum (Tables~\ref{tab:amplitude_range}
and~\ref{tab:fe_ranges}), consistent with regression to the mean under an
$L_2$ reconstruction loss. Extreme events are under-represented, which
matters for any downstream use involving intermittency.

\textbf{No divergence constraint.} Each velocity component is reconstructed
by a separate network, so nothing in the formulation enforces
$\nabla\!\cdot\!\mathbf{u} = 0$. The reconstructed field is not expected to
be divergence-free, and we have not quantified the residual. A joint
three-component model, or a divergence penalty added to
Eq.~\eqref{eq:loss}, would address this and is the most direct route to a
physically admissible output.

\textbf{Scope.} All results are for streamwise velocity in channel flow at
a single Reynolds number, on the mid-channel plane, from a single training
snapshot. Generalisation across Reynolds number, to other geometries, and
to the near-wall region where turbulence is most intense and anisotropic
is untested. The comparison is against non-learned interpolation only: no
learned baseline such as a 3D SRCNN or TEResNet was trained, so the
results establish that the model outperforms interpolation, not that it is
competitive with the current state of the art.

\section{Conclusion}

We presented a patch-based 3D-VAE for super-resolution of turbulent
channel flow whose parameter count is independent of the size of the domain
being reconstructed. On filtered DNS at $Re_\tau \approx 1000$ the model
reduces mean absolute error against DNS by $27\%$ relative to tricubic
interpolation and mean spectral amplitude error by approximately a factor
of three, and it transfers to coarse finite-element fields produced by a
different numerical scheme. A conditional 3D-GAN on the same data did not
converge; we report this and identify WGAN with a gradient penalty and
physics-based loss terms as the untried remedies.

The results should be read with the limitations of
Sec.~\ref{sec:limitations} in view. The 3D-VAE damps the smallest resolved
scales, introduces periodic artefacts at the patch stride, does not enforce
incompressibility, and has been evaluated on one velocity component at one
Reynolds number from one training snapshot against non-learned baselines.
What the experiments support is narrower than super-resolution in general:
a local, convolutionally applied coarse-to-fine operator recovers
energy-carrying and intermediate scales more faithfully than interpolation,
at a cost that does not grow with the domain.

\subsection{Future work}
\label{sec:future}

Four directions follow directly. First, the invariance the method relies on
is empirical: a locally applied convolutional operator implicitly assumes
statistical homogeneity of the coarse-to-fine relation, which is false near
walls. Conditioning on wall distance, or working in a local frame as
in~\cite{prakash2022invariant}, would replace an assumption with a
structure. Second, the sampling scheme should be tested at several Reynolds
numbers and against learned baselines to establish whether the operator is
genuinely method-agnostic and how it compares with current
architectures. Third, temporal consistency: the present test snapshot is
separated from the training snapshot by only $t^+ \approx 26$, and
evaluation at separations of several eddy turnovers is needed before any
claim of temporal generalisation can be made. Fourth, and most
substantially, the variational multiscale framework provides a principled
decomposition into resolved and unresolved scales with explicit coupling,
and the closure it requires is exactly a coarse-to-fine map of the kind
learned here. Embedding the learned operator in a variational multiscale
formulation, along the lines of Pradhan and
Duraisamy~\cite{pradhan2021variational}, would give the fine-scale
prediction a variational meaning it presently lacks.

\section*{Acknowledgment}

The author thanks the Johns Hopkins Turbulence Database team for providing
public access to the turbulent channel flow
dataset~\cite{lee2013petascale,graham2016web}, and the developers of
Oasis~\cite{mortensen2015oasis} for the finite-element solver.

\bibliographystyle{IEEEtran}
\bibliography{references}

\end{document}